\def\year{2022}\relax
\documentclass[letterpaper]{article} 
\usepackage{aaai22}  
\usepackage{times}  
\usepackage{helvet}  

\usepackage{comment}
\usepackage{placeins}
\usepackage{courier}  
\usepackage[hyphens]{url}  
\usepackage{graphicx} 
\usepackage{natbib}  
\usepackage{caption} 
\DeclareCaptionStyle{ruled}{labelfont=normalfont,labelsep=colon,strut=off} 
\usepackage{algorithm}
\usepackage{algorithmic}
\usepackage{amssymb}
\usepackage{enumitem}
\usepackage{multirow} 
\usepackage{tabularx}
\usepackage{zref-xr}
\usepackage[scientific-notation=true]{siunitx}
\usepackage{booktabs}
\usepackage{latexsym}
\usepackage{amsmath}
\usepackage{amsfonts}
\usepackage{xr}
\pdfoutput=1
\nocopyright
\makeatletter
\newcommand*{\addFileDependency}[1]{
	\typeout{(#1)}
	\@addtofilelist{#1}
	\IfFileExists{#1}{}{\typeout{No file #1.}}
}
\makeatother

\newcommand*{\myexternaldocument}[1]{%
	\externaldocument{#1}%
	\addFileDependency{#1.tex}%
	\addFileDependency{#1.aux}%
}

\myexternaldocument{supplimentary}

\usepackage{newfloat}
\usepackage{listings}
\floatstyle{ruled}
\newfloat{listing}{tb}{lst}{}
\floatname{listing}{Listing}
\title{Knowledge Graph Question Answering via SPARQL Silhouette Generation}

\author {
    Sukannya Purkayastha\footnote{Work done while the author was an intern at IBM Research},\textsuperscript{\rm 1}
    Saswati Dana, \textsuperscript{\rm 2}
    Dinesh Garg, \textsuperscript{\rm 2}
    Dinesh Khandelwal, \textsuperscript{\rm 2}
    G P Shrivatsa Bhargav \textsuperscript{\rm 2}
}
\affiliations {
    \textsuperscript{\rm 1} TCS Research\\
    \textsuperscript{\rm 2} IBM Research\\
    sukannya.purkayastha@tcs.com, sdana027@in.ibm.com, garg.dinesh@in.ibm.com, dikhand1@in.ibm.com, gpshri27@in.ibm.com
}

\usepackage{bibentry}
\usepackage{outlines}
\begin{document}
\maketitle
\pagenumbering{roman}
\setcounter{page}{1}

\begin{abstract}


Knowledge Graph Question Answering (KGQA) has become a  prominent area in natural language processing due to the emergence of large-scale Knowledge  Graphs  (KGs). Recently Neural Machine Translation based approaches are gaining momentum that translates natural language queries to structured query languages thereby solving the KGQA task. However, most of these methods struggle with out-of-vocabulary words where test entities and relations are not seen during training time. In this work,  we propose a modular two-stage neural architecture to solve the KGQA task.
The first stage generates a sketch of the target SPARQL called SPARQL silhouette for the input question. This comprises of (1) Noise simulator to facilitate out-of-vocabulary words and to reduce vocabulary size (2) seq2seq model for text to SPARQL silhouette generation.
The second stage is a Neural Graph Search Module. SPARQL silhouette generated in the first stage is distilled in the second stage by substituting precise relation in the predicted structure. We simulate ideal and realistic scenarios by designing a noise simulator. 
Experimental results show that the quality of generated SPARQL silhouette in the first stage is outstanding for the ideal scenarios but for realistic scenarios (i.e. noisy linker), the quality of the resulting SPARQL silhouette drops drastically. However,  our neural graph search module recovers it considerably.
We show that our method can achieve reasonable performance improving the state-of-art by a margin of $3.72 \%$ F1 for the LC-QuAD-1 dataset.
We believe, our proposed approach is novel and will lead to dynamic KGQA solutions that are suited for practical applications.

\end{abstract}

\section{Introduction}\label{sec:intro}

In recent years, there is an increasing interest in the Knowledge Graph Question Answering (KGQA)~\cite{diefenbach2018core} task in Natural Language Processing community due to its applicability in various real life and practical business applications. Availability of large-scale knowledge graphs, such as {Freebase}~\cite{bollacker2008freebase}, {DBpedia}~\cite{lehmann2015dbpedia}, {YAGO} \cite{10.1007/978-3-030-49461-2_34}, {NELL} \cite{NELL-aaai15}, and {Google's Knowledge Graph}~\cite{steiner2012adding} made this possible. The KGQA task requires a system to answer a natural language question leveraging facts present in a given KB. Mainly two streams of approaches are followed by KGQA community (1) semantic parse based (2) information extraction based. In Semantic parsed based approach, the task can be accomplished by translating the natural language question into a structured query languages or logic form such as SPARQL, SQL, $\lambda$-DCS~\cite{liang2013lambda}, CCG~\cite{zettlemoyer2005learning}, etc. Generated query is then executed over the given KG to finally arrive to the answer. Information extraction based approaches are primarily concerned with final answer but not intermediate logic form. In semantic parsed based approaches, main challenges in obtaining correct form of  logic/SPARQL is  getting the right structure along with specific entities and relations in the knowledge graph. 
Performance of existing off-the-shelf entity-relation linkers is not encouraging enough in KGQA dataset to adapt them in this task.  Therefore, most of the state-of-art systems follow Pipeline-based approaches with inbuilt entity-relation linker. These pipeline based approaches ~\cite{singh2018frankenstein,kapanipathi2020question,liang2021querying} are a popular way
to handle questions that requires multiple entities and relations to answer a given question~\cite{li2016dataset, usbeck20177th,trivedi2017lc}. The error introduced by inbuilt linkers is a major bottleneck and reduces the overall pipeline performance. 

\begin{figure*}[htb]
\centering
\includegraphics[width=2\columnwidth]{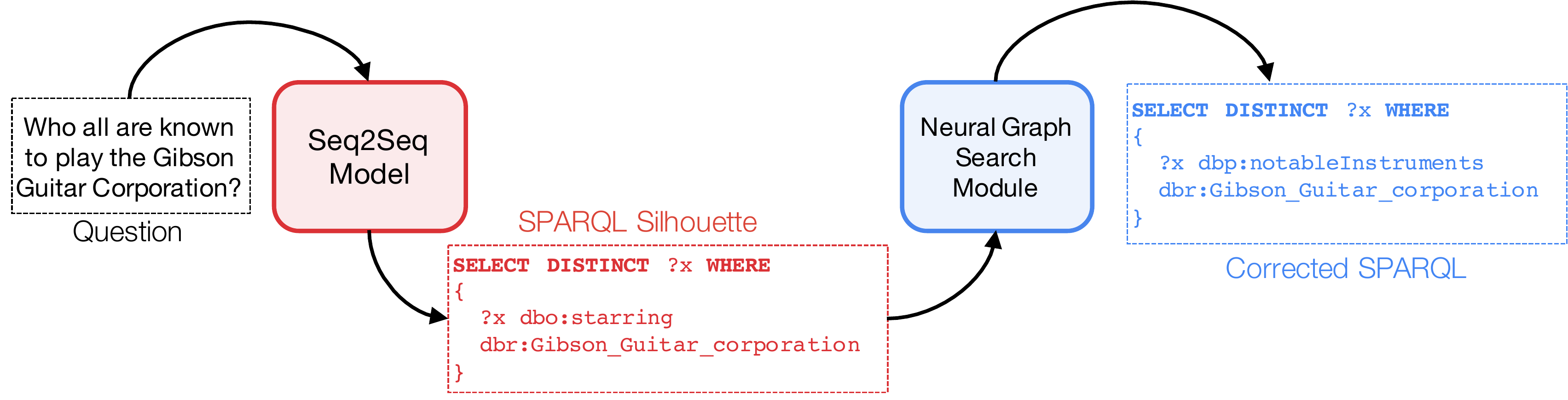}
\caption{A high level view of our proposed two stage neural architecture for KGQA.}
\label{two_stage_architecture}
\end{figure*}

With progress of neural network models, KGQA community aspires to perform the task by leveraging neural network. However to do so, we need large-scale training data which is a challenge.
These challenges limit the applicability of Deep Neural Network (DNN) based approaches on KGQA task. 
Existing neural approaches  however, are currently limited to answering questions that require single relation from KG~\cite{he2016character,DBLP:conf/acl/DaiLX16,hao2017end,lukovnikov2019pretrained,lukovnikov2017neural}. Some neural approaches ~\cite{maheshwari2019learning} assume a noise-free entity linker  or they mainly focus on relation linking sub-task ~\cite{yu2017improved}. ~\citet{hao2017end} follows information extraction based approaches and leverages universal KG information to arrive at the final answer more accurately.~\citet{cheng2018weakly} develops a system based on sequence-to-tree model where logic is in the latent form and supervision is in the form of final answer entity. Advances of translating natural language query to structured languages using NMT models ~\cite{yin2021neural,cai2017encoder} is emerging in recent years. In case of KGQA task, these NMT based methods suffer from out-of-vocabulary words and there is no explicit provision to handle unseen entities/relations at test-time.
However, we believe that the core challenges involved in performing KGQA task via NMT methods is not explored fully and there is a significant scope for further investigation. Motivated by these observations, in this paper, we propose a novel {\em two-stage neural architecture} (see Figure \ref{two_stage_architecture}) to answer KG based questions that need multiple entities and relations to answer them.

The main contributions of this work are as follows:
\begin{enumerate}[leftmargin=*,itemsep=0ex] 

\item 
In Stage-I, a sketch of {SPARQL} called {\em {SPARQL} silhouette} is generated for input question.
A noise simulator is designed in this module to devise three different kinds of masking strategies to simulate varying levels of noise introduced in entity/relation linking process.
\item 
In Stage-II, a simple and novel BERT based  {\em neural graph search module} (NGS) is proposed which corrects predicted relations in the SPARQL silhouette. Purpose of having this module is to overcome performance limitation that arises due to the weaknesses of entity/relation linker present in the first stage.
\item 
An ideal entity/relation linker having $100\%$ $F_1$ score  is simulated and shown that the quality of generated {SPARQL} silhouette is high --  
$83.08\%$ $F_1$ for {LC-QuAD-1} and $55.3\%$ Macro $F_1$ QALD for {QALD-9}. 
\item 
 Realistic scenario is simulated and shown that as $F_1$ of the linker goes down, quality of the resulting {SPARQL} silhouette drops drastically. Finally,  integrating Stage-II module with Stage-I boosts the performance significantly and improves the SOTA by a margin of $3.72\%$ $F_1$ for {LC-QuAD-1} dataset.
\end{enumerate} 

\section{Related Work}\label{sec:relatedwork}
In the beginning KGQA task was centralized in two directions either semantic parsed based ~\cite{unger2012template,berant2013semantic,reddy2014large,bast2015more,abujabal2017automated} approaches or information retrieval based ~\cite{bast2015more,yao2014information,dong2015question}.
Most of the earlier semantic parsed based approaches used handcrafted rules. We  limit our discussion only to end-to-end neural approaches.

\begin{table*}[htb]
\centering
\setlength\tabcolsep{4pt}
\renewcommand{\arraystretch}{1.1}
\begin{tabular}{lllll}
\toprule
\textbf{Approach} &
\textbf{Question}  & \centering  \textbf{Unseen}    & \textbf{Intermediate}  &  \textbf{Supervision}\\
& \textbf{Complexity}  & \centering  \textbf{Entities}    & \textbf{Logic Form}  & \\
\midrule
\cite{he2016character} & Simple & Yes  &  Yes &  Strong \\
\cite{yin-etal-2016-simple} & Simple &  Yes  &     No   &  Strong\\
\cite{hao2017end} & Complex	&	Yes	&	No	& Strong \\ 
\cite{maheshwari2019learning}& Complex &  No  &    Yes   &  Strong\\
\cite{cheng2018weakly} & Complex &  Yes  &     Yes  &  Weak \\
\cite{yin2021neural} & Complex &  No  &   Yes &   Strong \\
Our Approach & Complex & Yes  & Yes  & Strong\\
\bottomrule
\end{tabular}
\caption{Comparison of our approach with other neural network-based approaches. $2^{nd}$ column represents whether the approach can handle simple/complex questions. $3^{rd}$ column represents whether the approach can handle the unseen entities or not.  $4^{th}$ represents if the approach generates an intermediate logic form or not. $5^{th}$ column represents the type of supervision required to train the model; Strong means supervision using the manually annotated logical forms, whereas weak refers to supervision by providing only the correct denotations.}
\label{table:priorart}
\end{table*}
\subsection{Deep Neural Network Based Approaches }\label{section:NL2structedlanguage}
 With availability of large-scale datasets, DNN based techniques have made huge improvements in machine reading comprehension tasks~\cite{nguyen2016ms,rajpurkar2016squad,joshi2017triviaqa}. 
 This motivated NLP researchers to apply DNN technique to translate natural language question to structure database query languages ~\cite{Yu2018TypeSQLKT, wang2019rat,hosu2018natural,choi2021ryansql}.
 For KGQA, datasets like SimpleQuestions~\cite{bordes2015large, he2016character}, where only one entity and one relation are required to answer a question, performance of DNN models is already approaching the upper bound~\cite{petrochuk2018simplequestions}. To solve simple QA \citet{he2016character} use a char-level LSTM based encoder for the question and a char-level CNN to encode predicates/entities in KB. An attention based LSTM decoder is used to generate the topic entities and predicates. Whereas, to solve the complex KGQA task \cite{bao2016constraint, su2016generating, trivedi2017lc,dubey2019lc} it requires multiple KG facts. 
 To answer complex questions, \citet{hao2017end} first identify a topic entity from the question using FreeBase API and collect its 2-hop neighbours as potential answers. A cross-attention based Neural Net encodes the question w.r.t candidate answer aspects. Then they rank the candidates with similarity score based ranker without generating any intermediate logic form. Whereas, \citet{maheshwari2019learning} start with the gold entity in the question to generate the n-hop core chain of candidates. Then a bi-LSTM based slot matching model encodes the question and candidate core chains which are then ranked later. \citet{cheng2018weakly} use a bi-LSTM encoder and stack-LSTM decoder to generate logical forms with weak supervision.
 Recently
 NMT ~\cite{vaswani2017attention, bahdanau2014neural} based methods have also been used to solve KGQA task where a {\em seq2seq} model~\cite{he2016character, DBLP:conf/acl/DaiLX16,hao2017end,liang2017neural,cheng2018weakly} converts the natural language question directly into a logic form. \citet{yin2021neural} use CNN based seq2seq models to generate SPARQL queries from natural language questions. Their vocabulary for sparql generation is limited to the entities/relations seen during training. 
However, their performance reduces drastically if the overlap of entities and relations in the training and test sets differ as seq2seq models suffers from out-of-vocabulary words. 
Table \ref{table:priorart} shows the comparison of our approach with the neural network based previous approaches. To the best of our knowledge, our work is the first of its kind of solving KGQA task which considers multiple relations and used NMT method that handle out-of-vocabulary situation by designing noise simulator with masking strategy.

\section{The KGQA Task}\label{sec:problem}
In KGQA, we are given a Knowledge Graph $\mathcal{G}$ comprising of an entity set $\mathcal{E}$, a relation set $\mathcal{R}$, and a set of knowledge facts $\mathcal{F}$. The knowledge facts are expressed in the form of triples; $\mathcal{F} = \{\langle e_s,r,e_o\rangle\} \subseteq \mathcal{E} \times \mathcal{R} \times \mathcal{E}$, where $e_s \in \mathcal{E}$ is known as {\em subject} or {\em head} entity, $e_o \in \mathcal{E}$ is known as {\em object} or {\em tail} entity, and $r$ is a relation which connects these two entities. These entities (relations) form the nodes (edges) of the KG. 
The task now is to identify the subset of entities from $\mathcal{E}$ that constitute the answer of a given question $Q$ in the natural language form. The most common family of approaches for the KGQA task is {\em semantic parsing} where, the given question $Q$ is first translated into an {SPARQL} query $S$ which is then executed over the KG so as to get the answer set. 
For developing a system to convert a question into the corresponding {SPARQL} query, we are given a set of training data $\{Q_i,S_i, A_i\}_{i=1}^{n}$, where $Q_i$ is a question (in natural language text), $S_i$ is the {SPARQL} query, and $A_i$ is the answer set obtained by executing $S_i$ on $\mathcal{G}$. 
The proposed system consists of two stage neural modules. In the Stage-I, seq2seq module generates a SPARQL silhoutte with specific entities.  Relations predicted in this module are  corrected by the Stage-II, {\em neural graph search module}.

\section{Stage-I: Seq2Seq Model}\label{sec:stage1}
\begin{figure*}[t]
\centering
\includegraphics[width=\textwidth,height=3.5cm]{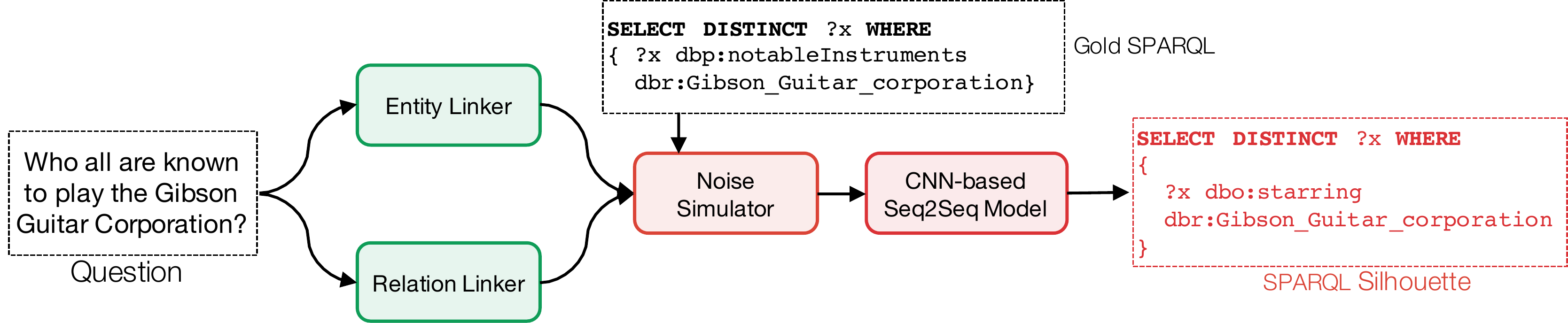}
\caption{A detailed architecture of Stage-I.}
\label{seq2seq_dia}
\end{figure*}

Sequence-to-sequence model have achieved state-of-the-art performance in machine translation ~\cite{yin2017syntactic} task. Encoder-decoder architecture of seq2seq models can vary from RNN, CNN based to transformer models. Prior research shows ~\cite{yin2021neural} that CNN based seq2seq model performs best among these for translating natural language to SPARQL query. Our preliminary experimental results were consistent with this fact since the CNN based model performed the best.
Hence, we moved ahead with the CNN based seq2seq model as our base model for Stage I. Figure \ref{seq2seq_dia} shows the architecture of Stage-I. An external entity/relation linker is used to detect surface form mentions of the entities/relations in the question text and linking the same to the underlying KG (DBpedia here). We designed a noise simulator for adapting the data to be in necessary format for seq2seq model. 

\subsection{Noise Simulator}
\begin{figure}[t]
\centering
\includegraphics[width=0.9\columnwidth,height=4.5cm]{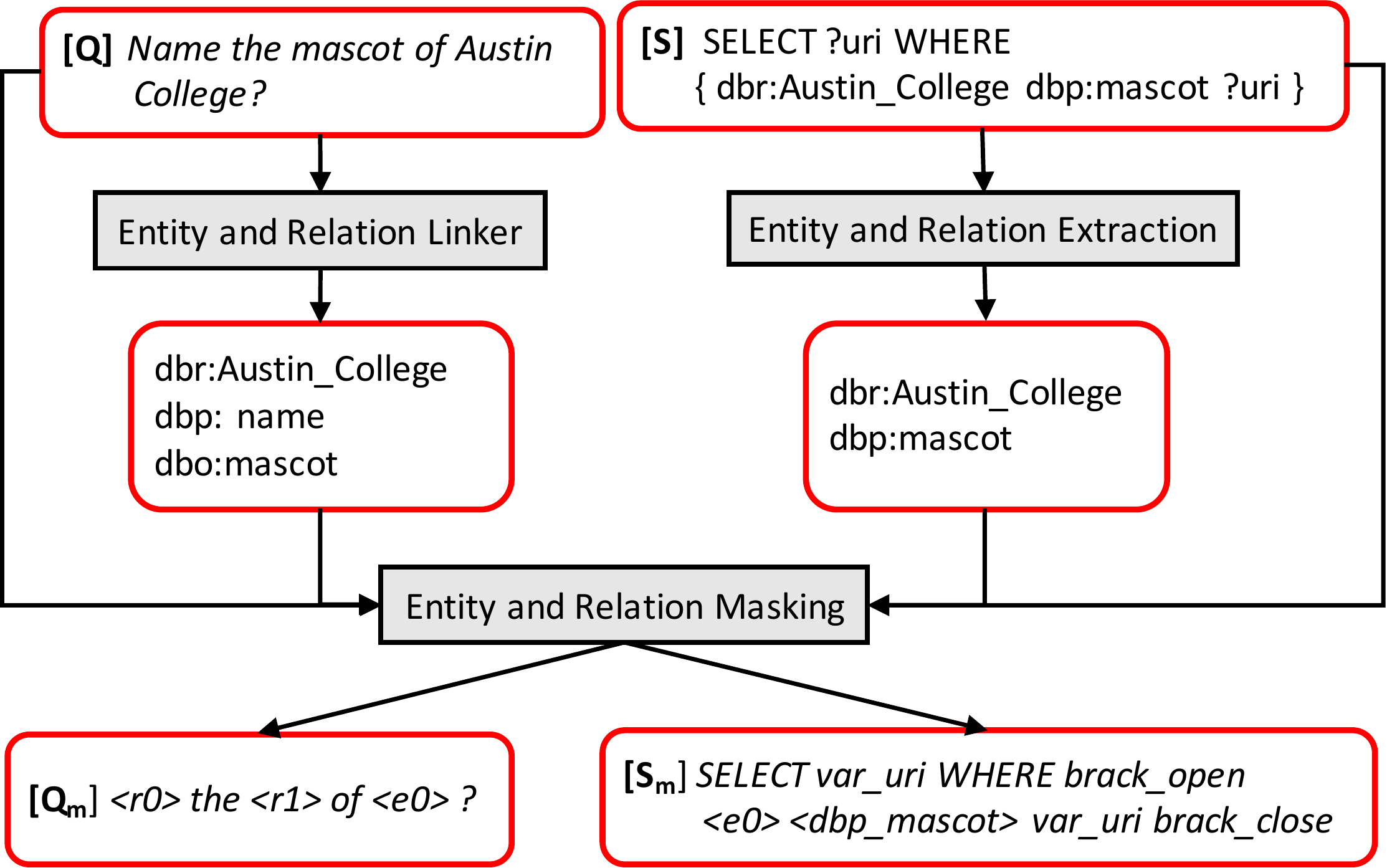}
\caption{An illustrative example for Scenario `C'. 
}
\label{mask_C}
\end{figure}

Purpose of designing noise simulator is twofold: (i) To simulate varying levels of noise in the entity/relation linking process (ii) To mask mentions and entities/relations in the question text and SPARQL.\\
{\bf [Need for Masking]} Masking helps in two ways: (1) handling test entities/relations that are unseen during training (2) reducing vocabulary size as KGs contain a large number of entities and relations. 
A simple neural seq2seq model which translates natural language question into a {SPARQL} query  will struggle to output some of the entities/relations during test time that are unseen during training time and hence will not be available in the output vocabulary. In the absence of linking and masking, our elementary experiments shows the performance of seq2seq model bo be very low with F1 score $ ~ 16 \%$ which was expected. 
\begin{table}[h]
\centering
\setlength\tabcolsep{4pt}
\renewcommand{\arraystretch}{0.9}
\begin{tabular}{llrr}
\toprule
Dataset & Statistics& Val & Test  \\
\midrule
\multirow{3}*{\parbox{2.0cm}{LC-QuAD-1}} & Entities (dbr) & $52.3$ & $46.8$ \\
& Properties (dbp) & $97.2$ & $98.3$ \\
& Ontologies (dbo) & $96.5$ & $94.6$ \\
\midrule
\multirow{3}*{\parbox{2.0cm}{QALD-9}} & Entities (dbr) & $27.1$ & $25.9$ \\
& Properties (dbp) & $0.0$ & $16.9$ \\
& Ontologies (dbo) & $47.8$ &$38.3$ \\
\bottomrule
\end{tabular}
\caption{\% of the entities and relations in val and test sets that are available within train set's gold {SPARQL}s.}
\label{table:coverage}
\end{table}
This outcome is obvious given the statistics in  Table \ref{table:coverage} which captures  percentage of entities and relations (i.e. properties and ontology in DBpedia) in validation and  test sets that are available in the training set.  This suggest that entity and relation linker is must for any neural model. Even if we use only neural models with perfect linkers, our SPARQL vocabulary dictionary will be over growing which becomes difficult to manage.
To handle the situation of increasing SPARQL vocabulary dictionary, we need masking/tagging techniques to mask entities and relations. \\
{\bf [Scenario `A': Noise-Free Linking]}
In this scenario, we simulate an entity and relation linker that has $100\%$ $F_1$. For this, we pick all entities/relations from the gold \texttt{SPARQL} and pretend as if they were the output of the linker (see Figure \ref{mask_A} in appendix). We begin with extracting  all the entities and relations from the gold {SPARQL} using their prefixes (\textit{dbr} for entities and \textit{dbp} or \textit{dbo} for relations). Next, we pick these entities and relations, and align the same with {\em surface-form mention text} in the given question. We observe that entities match exactly with substrings in the questions most of the time (e.g. {\em Austin College} in Figure \ref{mask_A} of the appendix). 
For relations, an exact match is not always possible, e.g., a given relation \texttt{dbo:film} is semantically best aligned to word {\em movies} in the question. We use pre-trained fastext embeddings \cite{bojanowski2017enriching} to represent words and relation and compute cosine similarity between each word in the question and the given relation. 
The highest-scoring word is considered as the aligned word. After identifying mentions of entities/relations, we mask them in question text and the corresponding gold SPARQL query. This masked pair is subsequently supplied to the seq2seq module as a training example.\\
{\bf [Scenario `B': Partly Noisy Linking]} 
Purpose of this scenario is to allow partial noise in the entity/relation linking process. 
For this, we first feed the natural language question into an external entity/relation linker. The linker returns two things: (i) A set of surface form mentions for entities/relations in the question text, and (ii) Linked entities/relations for these mentions. We take linker's output and find intersection of these entities/relations with the entities/relations present in the gold {SPARQL}. These common entities/relations are masked in the {SPARQL} query. Also, their corresponding surface forms are masked in the question text. In order to mask the surface forms in the question, we use exact match and string overlap based {\em Jaccard similarity}. Figure \ref{mask_B} in appendix illustrates this scenario.\\ 
{\bf [Scenario `C': Fully Noisy Linking]}
Goal here is to simulate a completely realistic scenario where we rely entirely on an external entity/relation linker. 
For this, we feed input question to the entity/relation linker and get the suggested surface form mentions and linked entities/relations. We mask each of these suggested mentions using exact match and partial match. Corresponding {SPARQL} query's entities and relations are also masked based on the suggestions. This scenario is depicted in Figure \ref{mask_C}.
\subsection{Convolutional \texttt{Seq2Seq} Model}\label{sec:problem}
\begin{figure*}[t]
	\centering
	\begin{minipage}{.65\textwidth}
		\centering
		\includegraphics[width=.95\linewidth,height = 10cm]{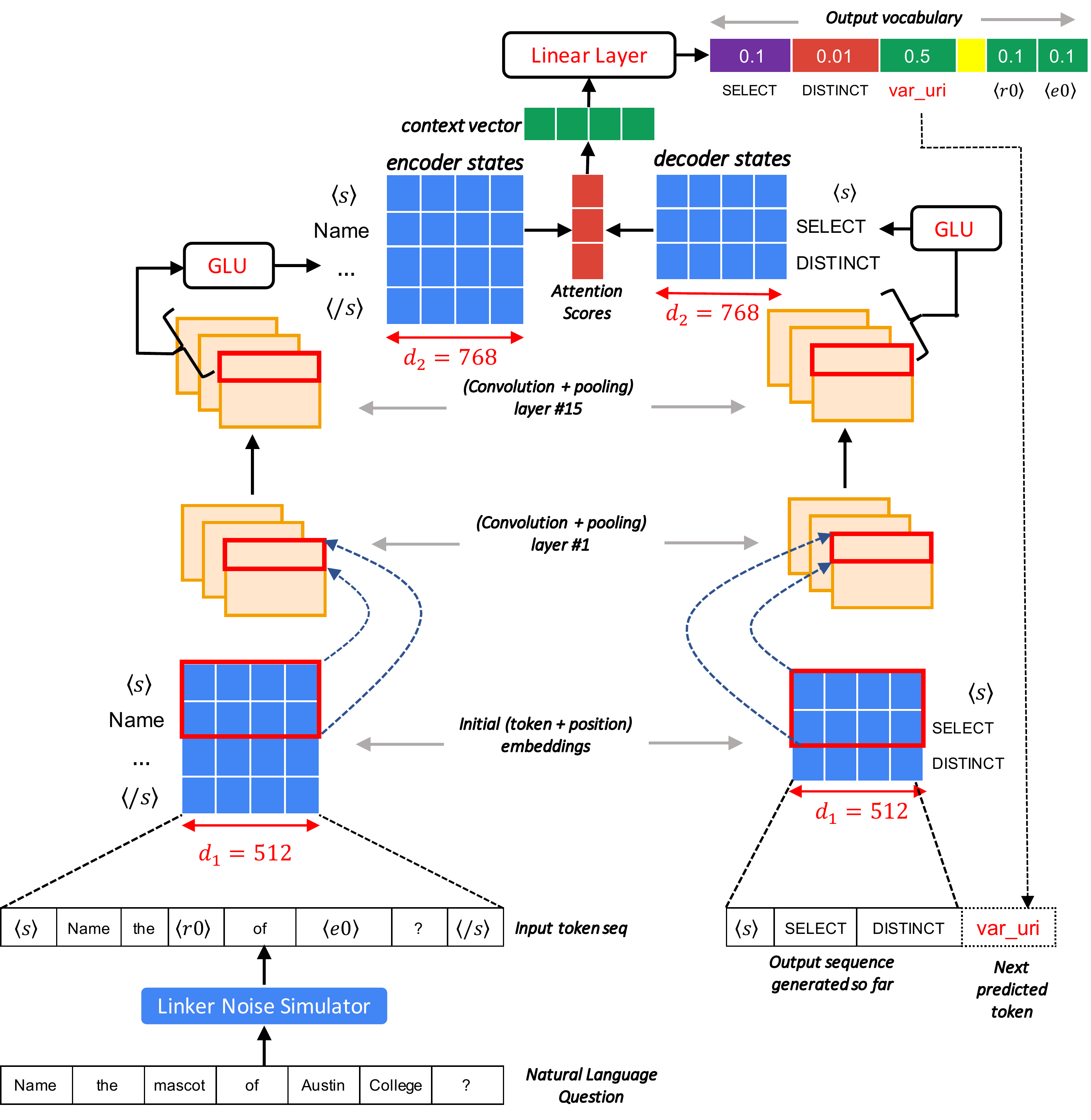}
		\captionof{figure}{A CNN-based \texttt{Seq2Seq} model for KGQA. We have assumed noise-free linking scenario here.}
		\label{cnn_archi}
	\end{minipage}%
	\begin{minipage}{.35\textwidth}
		\centering
		\includegraphics[width=.95\linewidth]{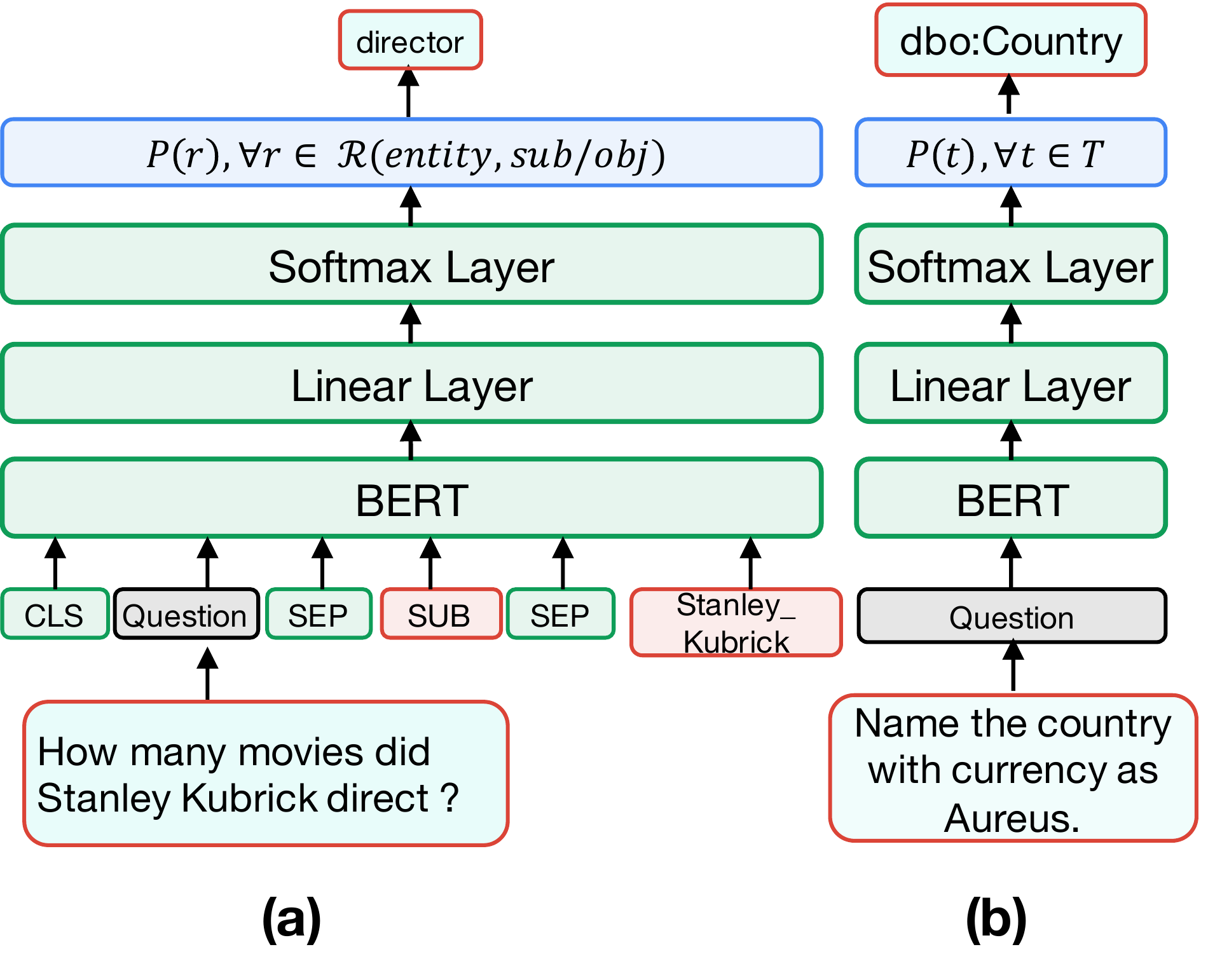}
		\captionof{figure}{Architecture of neural graph search module. (a) Relation Classifier. This module predicts relation for a given entity (b) Ontology Type Classifier. This module predicts \textit{rdf:type} ontology class.}
		\label{archi}
	\end{minipage}
\end{figure*}

The pair of masked question and SPARQL query obtained from the noise simulator, under any noise scenario, is fed to a  {\em Convolutional Neural Network (CNN)} based {\em seq2seq} model~\cite{gehring2017convolutional}. As shown in Figure \ref{cnn_archi}, this model reads the entire masked question and then predicts the corresponding masked SPARQL query token-by-token in a left-to-right manner. This seq2seq model consists of the following key components. 
\\{\bf [Input Embedding Layer]}
Both encoder and decoder consist of an embedding layer that maps each input token to a point-wise summation of its word embedding and positional embedding. The embedding of each word is initialized randomly. In order to capture the sense of order, the model is provisioned with the positional embedding.\\ 
{\bf [Convolution + Pooling Layers]} The token embeddings obtained from the previous layer are fed to the multiple convolution and pooling layers. Each convolution layer consists of a $1$-dimensional convolution followed by Gated Linear Units (GLU) \cite{dauphin2017language}. 
Residual connections \cite{he2016deep} are added from input to the output of each convolution layer.\\ 
{\bf [Multi-Step Attention]} Each decoder layer comprises a convolution layer followed by a multi-step attention layer. This multi-step attention is used to find the attention scores from a particular decoder state to the source tokens. 
Attention between 
decoder state $d_i$ (after $i^{th}$ layer) of the last token in  generated sequence so far and state $z_j$ of the $j^{th}$ source element (after last encoder layer) is computed as: $a_{j}^{i}={exp(d_{i}\cdot z_j)}/{\sum\nolimits_{t=1}^{m}exp(d_{i} \cdot z_{t})}$
where, $m$ is the number of source elements.
The context vector, $c_i$, is now computed as, $c_i = [\sum_{j=1}^{m} a_{j}^{i}(z_j+e_j)]+d_i$ where, $e_j$ is the input embedding for the source element $j$.\\ 
{\bf [Output Layer]} Finally, output at a particular time step is calculated over all the $Z$ possible tokens, $P(z_{t+1}|z_1,\dots , z_t,X) = softmax(Wd_L+b)$
where $P(z_{t+1}|\cdot) \in \mathbb{R}^{Z}$, and $W$, $b$ are trainable parameters. $d_L$ is the decoder state of last target element at the last layer $L$. $X$ is the input sequence.\\
{\bf [Training Loss:]}
The model is trained using {\em label smoothed cross-entropy loss} given by following expression (for single training example) $L(\theta) = -({1}/{N})\cdot\sum\nolimits_{n=1}^{N}\sum\nolimits_{z=1}^Z q(y_n=z|y_{n-1}) \cdot \log P_{\theta}(y_n=z|y_{n-1})$ where, $N$ is the number of words in output sequence and $y_n$ is the first $n$ tokens of output sequence. $P_{\theta}(y_n=z|y_{n-1})$ is model's probability to output token $z$ given $y_{n-1}$ sequence generated so far. The quantity $q(y_n=z|y_{n-1})$ is equal to $\gamma$ if $f(y_n)=z$ and ${(1-\gamma)}/{(Z-1)}$ o/w, where
$\gamma \in [0,1]$, $\gamma >{1}/{Z}$.
\section{Stage-II: Neural Graph Search Module} 
While working with LC-QuAD-1 and QALD-9 datasets, our error analysis on output of Stage-I revealed that entity linking performance is reasonably good but the same is not true for relation linking. Existing literature ~\cite{wu-etal-2020-scalable, Li2020EfficientOE} also show enough evidences of achieving high performance on the entity linking task, whereas relation linking turns out to be harder due to complexity of natural language. 
Because of this, we have most of the entities within a {SPARQL} silhouette generated by Stage-I as correct but the relations are incorrect. Graph search module in Stage-II takes a {SPARQL} silhouette as input and produces an improved version of the same by replacing incorrect relations(See Figure \ref{input_graph_search} in appendix for an example).\footnote{It is easy to extend this idea and perform an iterative graph searching when entity linker performance is also low.} This is a BERT-based module and its architecture is shown in Figure \ref{archi}. This module works as follows. 
\begin{enumerate}[leftmargin=*,itemsep=0ex]
\item One-by-one, we consider each triple $\langle e_s, r, e_o\rangle$ in the {SPARQL} silhouette and try correcting its relation $r$ through this module. Note, in triple $\langle e_s, r, e_o\rangle$, at least one of the entity must be an existential variable unless it is an {\em rdf:type} relation, which we handle separately. We consider this triple for the correction only if the other entity is grounded to some KB entity and that grounded entity could be in subject (or object) position.
\item For each such triple identified in the previous step, we prepare input in the following format:
[CLS] $Q$ [SEP] [SUB (or OBJ)] [SEP] $e_s$ (or $e_o$). Here, $Q$ is token sequence of input question text and [SUB (or OBJ)] is special token depending on whether the grounded entity is in subject (or object) position (refer Figure \ref{archi}a). We also pass grounded entity ($e_s$ or $e_o$) as the last element of this input. [CLS] and [SEP] are special tokens from BERT vocabulary. 
\item We feed above input sequence of tokens into the BERT layer of graph search module. The output is passed through a linear layer followed by a {\em softmax} layer. This softmax layer induces a probability score $p_r$ for each relation $r\in \mathcal{R}$ in the given KG. 
While training, we use the following loss function (given for single example): $\ell= (1-\alpha)* (\ell_c) + (\alpha) * (\ell_{gs})$. Here, $\ell_c$ denotes standard cross entropy loss between predicted probabilities $\{p_r\}_{r \in \mathcal{R}}$ and the gold relation. The graph search loss term $\ell_{gs}$ forces the predicted probabilities to be low for all those relations which are invalid relations (in the given KG) for corresponding input entity $e_s$ (or $e_o$) in the input position (subject or object). For this, we assume a uniform probability distribution over all such valid relations and compute its cross entropy loss with $\{p_r\}_{r \in \mathcal{R}}$.  $\alpha$ is a hyperparameter.
\item During inference, at softmax layer, we restrict the outputs only to those relations $r\in\mathcal{R}$ which are valid relation for the input entity as being subject or object. For example, if input grounded entity is $e_s$ then we restrict prediction to only those relations $r$ for which $\langle e_s, r, ?x\rangle$ is a valid triple for some grounding of $?x$. In DBpedia same relation can exist in the form of \textit{`dbo'} and \textit{`dbp'} for a specific entity. 
In such cases, we always pick the \textit{`dbo'} version.
Prediction is  made based out of $61623$ relations available in DBpedia.
\item If a relation $r$ in a given triple is {\em rdf:type} then we handle them little differently. Note, in DBpedia, a triple containing \textit{rdf:type} relation looks like this $\langle ?x, \textit{rdf:type}, \textit{dbo:type} \rangle$ where, $?x$ is a variable and \textit{dbo:type} is the DBpedia ontology class of the entity $?x$. For such triples, we maintain a separate version of the neural graph module (refer Figure \ref{archi}b). Input to this module is [CLS] Q. We need to predict the corresponding ontology class \textit{dbo:type}. DBpedia ontology contains $761$ classes and hence, in this model, prediction is one of these $761$ classes. This module is trained with standard cross-entropy loss. An example of the \textit{rdf:type} classification would be to predict \textit{dbo:Country} for the question \textit{`Name the country with currency as Aureus?'}.
\end{enumerate}



\section{Experiments}\label{sec:experiments}
\paragraph{Datasets:}
We work with two different KGQA datasets based on DBpedia: {LC-QuAD-1}~\cite{trivedi2017lc} and {QALD-9}~\cite{ngomo20189th}. 
{LC-QuAD-1} contains $5000$ examples and is based on the {\em 04-2016 version} of the {DBpedia}. We split this dataset into $70\%$ training, $10\%$ validation, and $20\%$ test sets (same as the leaderboard). {QALD-9} is a multilingual dataset and is based on the {\em 10-2016 version} of the {DBpedia}. Questions in this dataset vary in terms of reasoning nature (e.g. counting, temporal, superlative, comparative, etc.) and therefore, in terms of the {SPARQL} aggregation functions as well. This dataset contains $408$ training and $150$ test examples. We split the training set into $90\%$ training and $10\%$ validation sets. 
\paragraph{Evaluation Metric:}
Performance is evaluated based on the standard precision, recall, $F_1$ score for KGQA systems. For more detail please refer to \ref{eval_met}.\\ 
{\bf Baseline:} We compare our approach with three baselines: WDAqua ~\cite{Diefenbach2020TowardsAQ}, QAmp ~\cite{vakulenko2019message} and gAnswer \cite{zou2014natural}. WDAqua is a graph based approach where authors first develop SPARQL query based on four predefined patterns. In the second step they rerank generated candidates. QAmp used text similarity and graph structure  based on an unsupervised message-passing algorithm. gAnswer is graph data driven approach and generate query graph to represent user intention.

\paragraph{Experimental Setup:}\parbox{0.1cm}{ }\\
{\em 1) Stage-I:} We use {\em Falcon}~\cite{sakor2019falcon} for entity/relation linking and experiment with all $3$ noise scenarios. 
We use {\em fairseq}\footnote{https://github.com/pytorch/fairseq} library for implementation of CNN based seq2seq model~\cite{gehring2017convolutional} comprising of $15$ layers\footnote{We will release our code after the review period.}. and used Nesterov Accelerated Gradient (NAG) optimizer. 
We experimented with different values of hyperparameters and report results for the values yielding the best performance on the validation set. 
Details about tuning ranges and optimal values of all these hyperparameters are given in Table \ref{table:hyper-param} and Figure \ref{fig:val_accuracy} of appendix. We used $2$ Tesla v100 GPUs for training seq2seq model.
\\
{\em 2) Stage-II:} For neural graph search module, we work with a pre-trained {\em BERT-base uncased model}. It consists of $12$ transformer layers, $12$ self-attention heads, and $768$ hidden dimension. We used $1$ Tesla v100 GPU for training.

\section{Results}\label{sec:results}
Table \ref{table:lcquad} compares the performance of our model with baseline models for the LC-QuAD-1 dataset. The first two rows are top entries in the LC-QuAD-1 leaderboard \footnote{http://lc-quad.sda.tech/lcquad1.0.html}. 
The next set of rows show result of our approach. Our results of stage-II are under realistic scenario or full noise  setting for entity/relation linking.

\begin{table}[t]
\centering
\setlength\tabcolsep{2.5pt}
\renewcommand{\arraystretch}{0.9}
\begin{tabular}{llrrrr}
\toprule
\parbox{0.5cm}{\centering Model Type} & \parbox{0.5cm}{\centering Model Name} &  AM   &   Prec.  &    Recall  & $F_1$  \\
\midrule
\multirow{2}*{Baseline} & WDAqua & - & $22.00$ & $38.00$ & $28.00$ \\ 
& QAmp  &   -       &   $25.00$    &   ${\bf 50.00}$    &   $33.33$      \\
\midrule
\multirow{3}*{\parbox{1.2cm}{\centering Stage-I (Ours)}} & No Noise	&	 $82.88$	&	 $83.11$	&	 $83.04$	&	 $83.08 $\\
& Part Noise   & 	 $41.34$	&    $42.40$   &   $42.26$   &   $42.33$ \\
& Full Noise    & $24.92$   &    ${\bf 25.54}$   &    $25.64$   &    $25.59$ \\
\midrule
\multirow{2}*{\parbox{1.3cm}{\centering Stage-II (Ours)}} & w/o type	& $30.63$ & ${\bf 32.17 }$ & $32.20$ & $32.18$\\
&w/ type    & 
$34.83$   &    ${\bf 37.03}$   &    $37.06$   &    ${\bf 37.05}$ \\
\bottomrule
\end{tabular}
\caption{Test set performance on {LC-QuAD-1} dataset. 
 }
\label{table:lcquad}
\end{table}
Table \ref{table:qald} captures the performance of our approach on QALD-9 dataset. The first two rows in Table \ref{table:qald} correspond to a baseline model and a top entry in the {QALD-9} challenge ~\cite{ngomo20189th}. The next set of rows show performance of our model.
\begin{table}[t]
\centering
\setlength\tabcolsep{2pt}
\renewcommand{\arraystretch}{1.2}
\begin{tabular}{llllllr}
\toprule
\parbox{0.5cm}{\centering Model Type} & \parbox{0.5cm}{\centering Model Name} &  \parbox{0.8cm}{\centering AM}  & \parbox{0.8cm}{\centering Mac. Prec.}  & \parbox{0.8cm}{\centering Mac. Rec.}  & 
\parbox{0.5cm}{\centering Mac. $F_1$} & 
\parbox{0.8cm}{\centering Mac. $F_1$ QALD}  \\
\midrule
\multirow{2}*{Baseline}
& WDAqua & - & $26.1$ & $26.7$ & $25.0$ &
$28.9$ \\
& gAnswer & - & $29.3$ & ${\bf32.7}$ & $29.8$ & ${\bf43.0}$ \\  
\midrule
\multirow{3}*{\parbox{1.2cm}{\centering Stage-I (Ours)}} &No Noise &	$29.9$ &	 $80.4$	&  $42.1$ &	$40.9$ & $55.3$	\\
&Part Noise &  $13.1$ &  $63.9$ &  $28.7$ & $22.4$ & $39.6$ \\
&Full Noise &  $11.1$ &  ${\bf82.6}$ &  $23.0$ & $20.6$ & $36.0$ \\   
\midrule
\multirow{2}*{\parbox{1.3cm}{\centering Stage-II (Ours)}} & w/o type	& $15.3$ & ${\bf59.4}$ & $26.1$ & $23.3$ & $36.2$\\
&w/ type    &  $15.3$ & ${\bf59.4}$ & $26.1$ & $23.3$ & $36.2$\\
\bottomrule
\end{tabular}
\caption{Test set performance on \texttt{QALD-9} dataset. Here Mac. means Macro and Rec. means Recall.
}
\label{table:qald}
\end{table}

\begin{figure}
\centering
\includegraphics[width=6.0 cm,height=4.5cm]{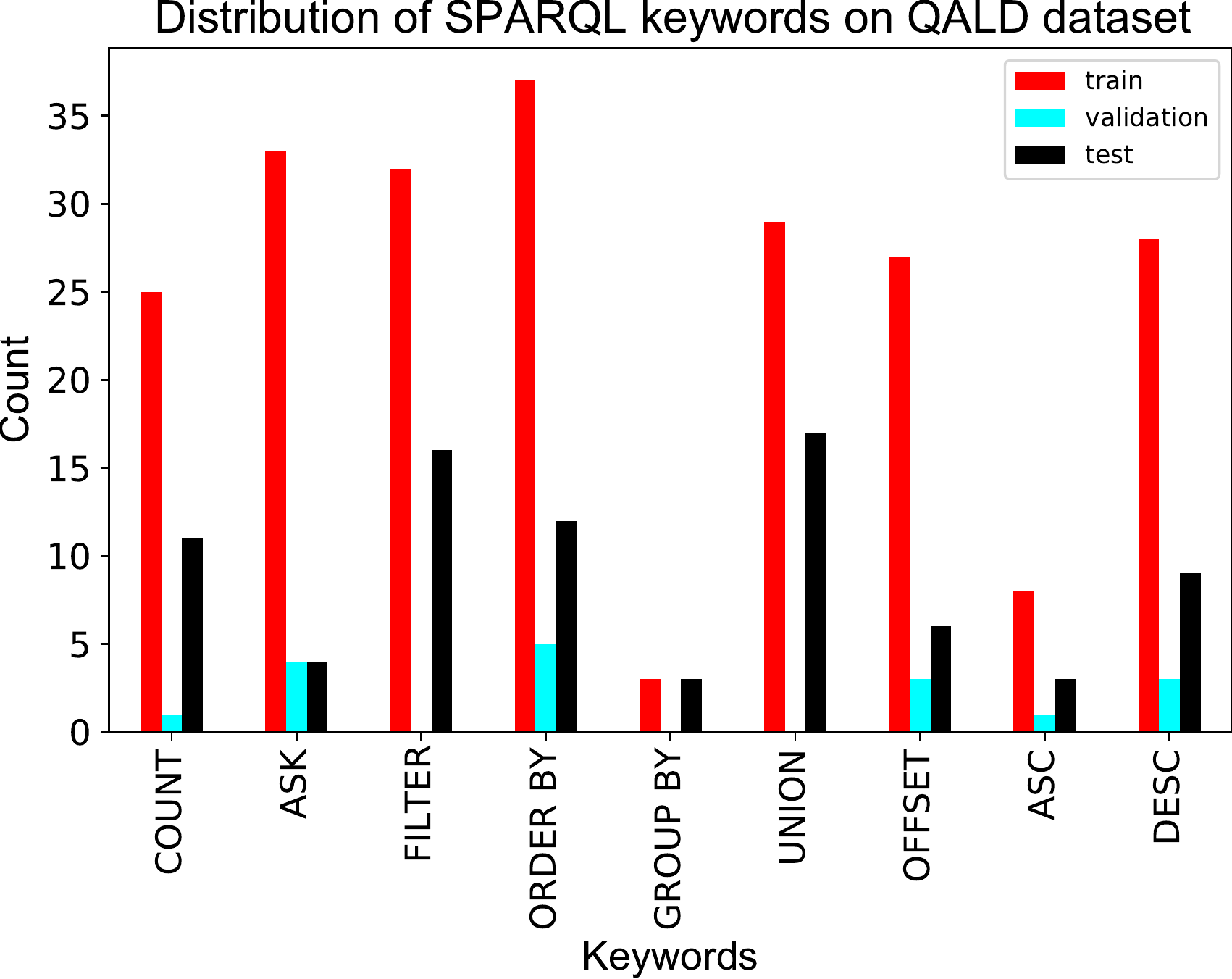}
\caption{Frequency of {SPARQL} keywords in {QALD-9}.}
\label{keyword_dis}
\end{figure}

{\bf Insights:} From Tables \ref{table:lcquad} and \ref{table:qald}, one can observe that performance of Stage-I under {\em No Noise} linking becomes an upper bound on the performance of seq2seq model. This means seq2seq model can achieve upto $83.08 \%$ $F_1$ for LC-QuAD-1 and $55.3\%$ Macro $F_1$ QALD for QALD-9 dataset if the entity/relation linker were to be $100\%$ correct. The gap between the performance of {\em No Noise} linking (upper bound) and {\em Full Noise} linking (lower bound) illustrates how the performance of entity/relation linker impacts the overall performance of KGQA. 
Further, the performance of Stage-II demonstrates how one can improve the lower bound numbers by adopting our proposed graph search module. For LC-QuAD-1, we gain $11.46\%$ in $F_1$ in Stage-II whereas, for QALD-9 this gain is only $0.2\%$ in Macro $F_1$ QALD. For QALD-9 dataset, the numbers in last two rows are same because we have only two questions with \textit{rdf:type} and their classes belong to {\em YAGO} ontology so our model does not predict them. One may also observe that the overall performance after Stage-II improves the respective baseline in case of LC-QuAD-1 dataset but QALD-9 dataset it struggles.\\  
{\bf Error Analysis:} Reason for QALD-9 having low upper bound is its training set size being too small ($367$).
 Further analysis reveals that QALD-9 dataset has large variety of SPARQL keywords from a small train set. Figure \ref{keyword_dis} captures the distribution of SPARQL keywords in QALD-9 dataset (excluding SELECT and DISTINCT keywords as they appear in almost all the questions). From this figure, its clear that number of questions varies from $3$ to $37$ for each category of SPARQL keywords which is too less for any neural model to learn from. We also trained our model with combining LC-QuAD-1 and QALD-9 dataset in both the stages. But it did not improve the performance of QALD-9 dataset because the nature of the SPARQL is very different in both the datasets.
Because of these reasons, unlike LC-QuAD-1, the generated {SPARQL} silhouette in QALD-9 dataset has errors other than incorrect entity/relation. Therefore, Stage-II offers much smaller gain for QALD-9 relative to LC-QuAD-1.
Lastly, Table \ref{table:falcon} which demonstrates the performance of Falcon linker on test set for both the datasets rules out the possibility of systematic data bias in terms of entity/relation linking. Though entity linking performance is reasonable for both the datasets, relation linking is consistently substandard for both the datasets. The poor performance of Falcon on relation linking also justifies a substantial gap between upper and lower bounds for both datasets.

\begin{table}[t]
\centering
\setlength\tabcolsep{3.0pt}
\renewcommand{\arraystretch}{0.9}
\begin{tabular}{lccc}
\toprule
Dataset &  E/R  &  Precision (\%)  & Recall (\%)  \\
\midrule
\multirow{2}*{\parbox{2.6cm}{LC-QuAD-1}} &
E & $79.19$  &   $85.60$     \\
 & R       & $43.74$   &    $44.99$   \\
\midrule
\multirow{2}*{\parbox{1.6cm}{QALD-9}} & E & $78.00$  &   $98.55$     \\ 
& R       & $41.05$   &    $37.17$   \\
\bottomrule
\end{tabular}
\caption{{\em Falcon} performance on entity (E) and relation (R) linking on test sets.}
\label{table:falcon}
\end{table}
{\bf Anecdotal Examples:} 
Table \ref{failing_examples} of appendix shows examples from LC-QuAD-1 test set where our neural graph search module struggles to disambiguate between two very similar looking relations that exist in DBpedia for an entity. Table \ref{atypical_examples} captures examples from QALD-9 test set where gold {SPARQL} have an intrinsic structure because the way in which corresponding facts are being captured within DBpedia. This makes it difficult for any KB agnostic techniques (such as seq2seq) to output such structures. Finally, Table \ref{missing_for_less_freq_keyword} shows examples from QALD-9 test set where gold SPARQL comprises infrequent SPARQL keywords making it hard for seq2seq model to learn about them.


\section{Conclusions}

We propose a simple sequential two-stage purely neural approach to solve the KGQA task. We demonstrate that, if entity/relation linking tasks are done perfectly, then Stage-I, vanilla seq2seq neural module can produce impressive performance on KGQA task.
However, in noisy realistic scenarios, it performs differently. We have proposed a novel Stage-II, a {\em neural graph search} module to overcome noise introduced by entity/relation modules. Our approach improves state-of-art performance for LC-QuAD-1 dataset. Though, for QALD-9 dataset due to the small training size and intrinsic nature of facts in DBpedia, our model struggles to improve state-of-art, we believe, this research demonstrates great potential of pure neural approaches to solve the KGQA task and opens up a new research direction.

\bibliography{main}

\newpage
\section{Appendix}

\section{Noise Simulator}
\begin{figure}[h!]
\centering
\includegraphics[width=0.7\columnwidth]{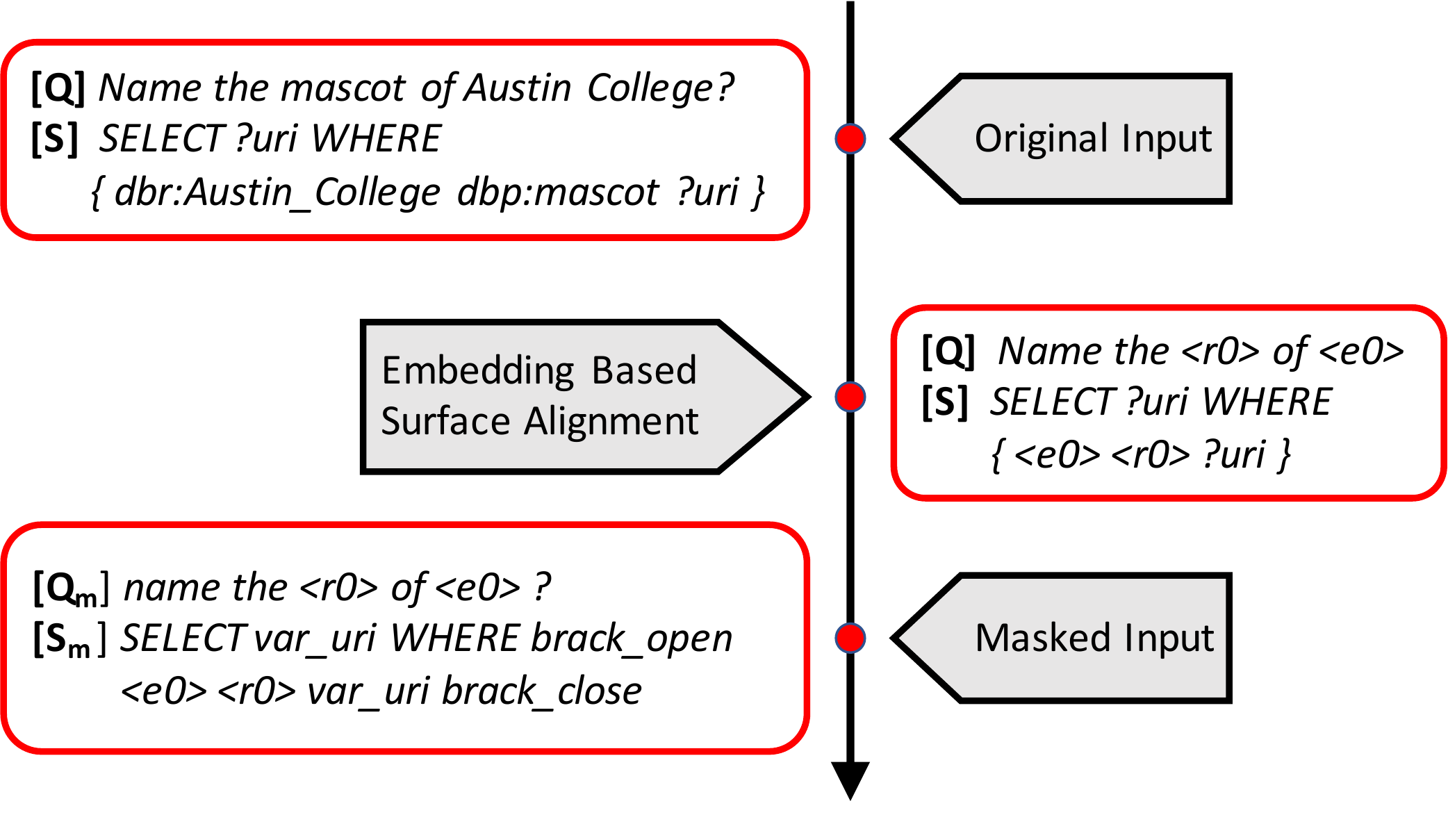}
\caption{An illustrative example for Scenario `A': Noise-Free Linking. To align the surface forms of the entities/relations mentions in the given question text, we used word embedding as it offers higher alignment $F_1$. We used Falcon as a linker.} 
\label{mask_A}
\end{figure}
\begin{figure}[h!]
\centering
\includegraphics[width=0.9\columnwidth]{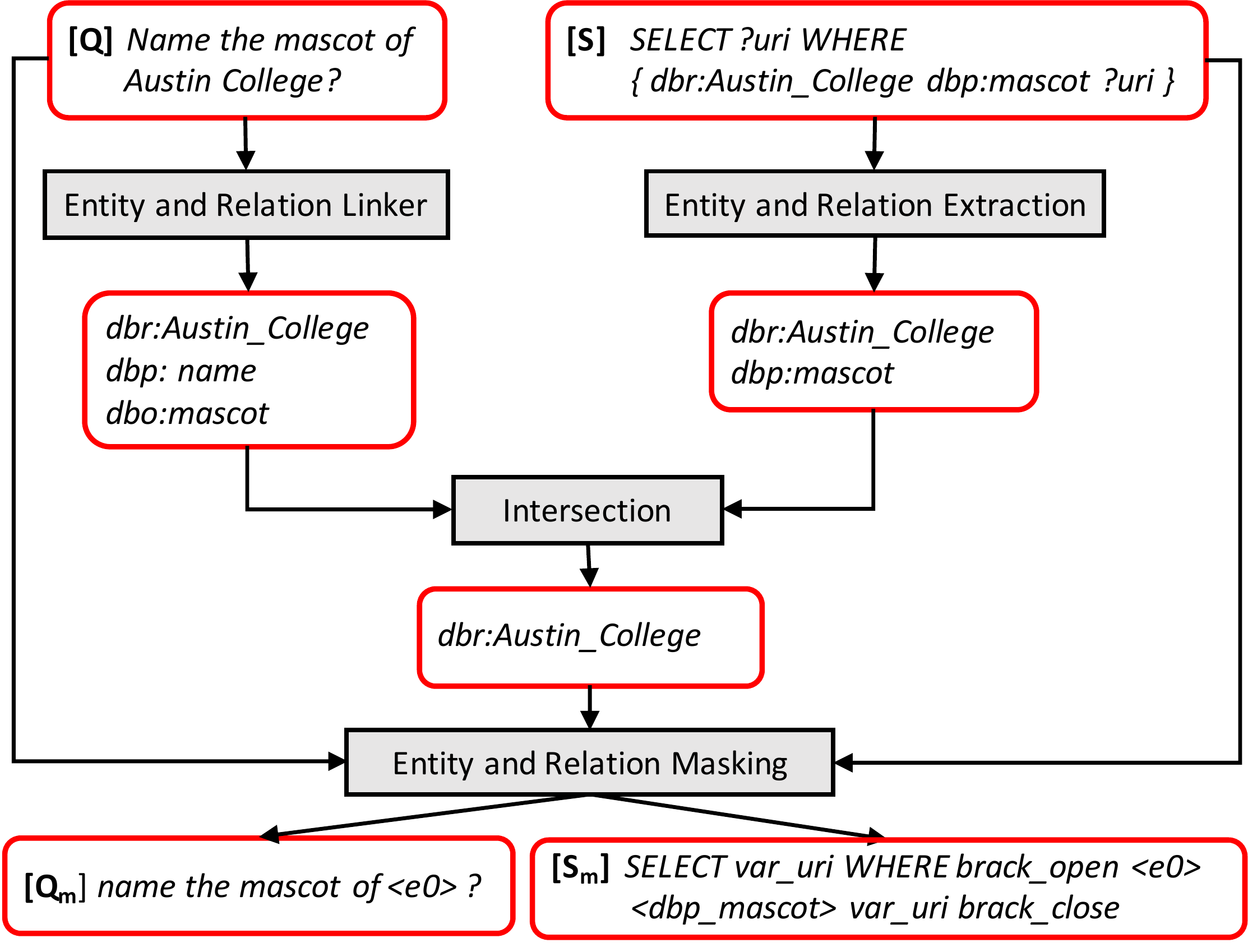}
\caption{An illustrative example for Scenario `B': Partly Noisy Linking. To align the surface forms of the entities/relations mentions in the given question text, we used exact match as well as string overlap based {\em Jaccard similarity} with a threshold of $0.7$. We used Falcon as a linker.} 
\label{mask_B}
\end{figure}


\begin{table*}[h!]
\centering
\setlength\tabcolsep{2pt}
\renewcommand{\arraystretch}{1.4}
\begin{tabular}{lrr}
\toprule
Hyperparameter &\parbox{2cm}{\centering Tuning Range} & \parbox{1cm}{\centering Best Value} \\
\midrule
$\eta$ for Stage-I        & [\num{0.1}, \num{0.2}, \num{0.25}, \num{0.50}] & $0.25$\\
$\eta$ for Stage-II  & [$10^{-4}$, $10^{-5}$, $10^{-6}$]           &   $10^{-5}$    \\
$b$ for both stages          & \num{8}                                        & \num{8} \\
$\alpha$ for LC-QuAD-1          & [\num{0.1}, \num{0.4}, \num{0.6}, \num{0.7}] & \num{0.4}  \\
$\alpha$ for QALD-9             & [\num{0.1}, \num{0.4}, \num{0.6}, \num{0.7}] & \num{0.6}      \\
\bottomrule
\end{tabular}
\caption{Tuning range and the final chosen best values of various hyperparameters. $\eta$ means learning rate and $b$ means batch size.}
\label{table:hyper-param}
\end{table*}
\begin{figure*}[h!]
\centering
\includegraphics[width=1.5\columnwidth]{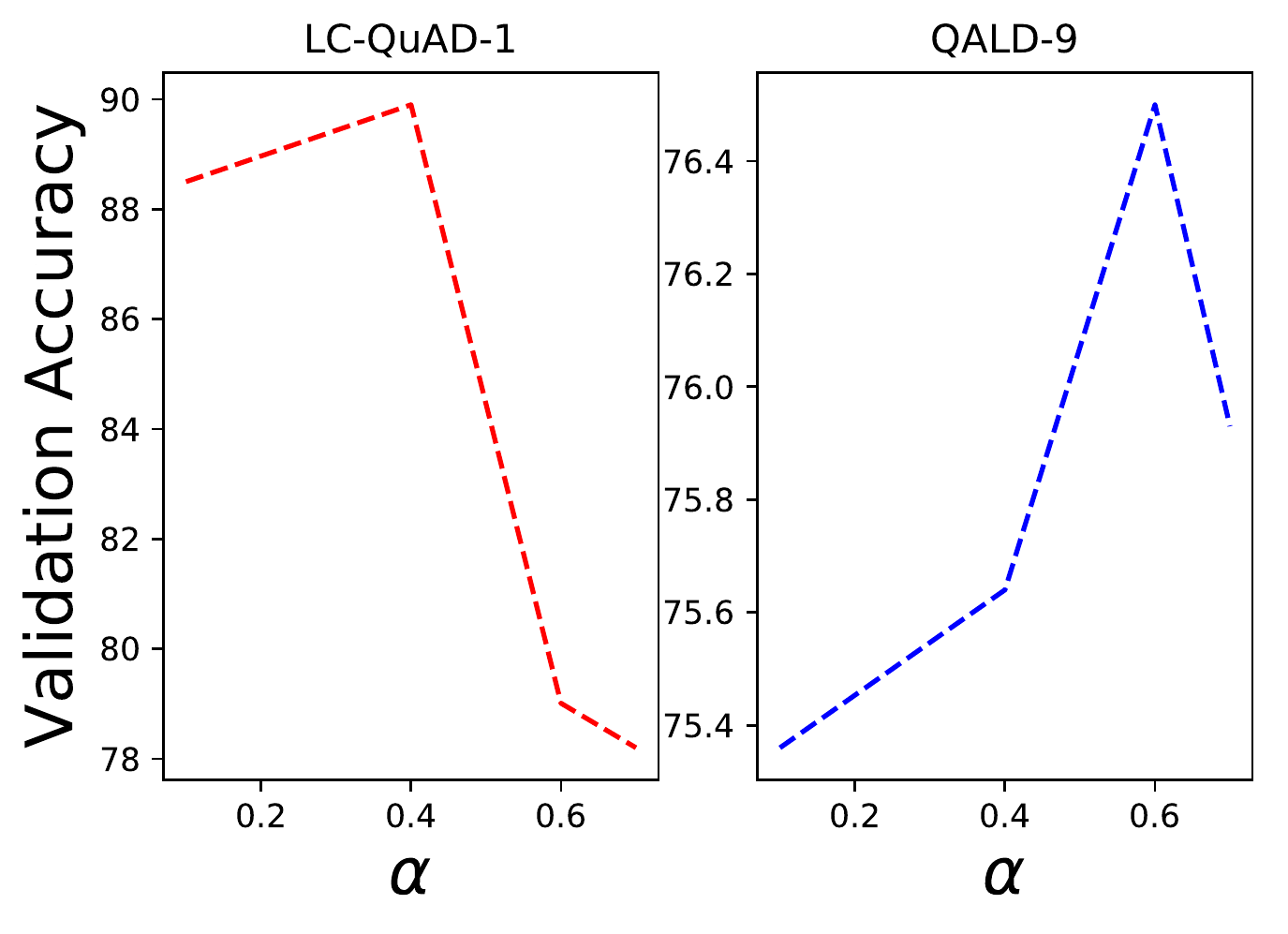}
\caption{Change in validation set accuracy with hyperparameter $\alpha$}
\label{fig:val_accuracy}
\end{figure*}

\begin{figure*}[h!]
\centering
\includegraphics[width=2\columnwidth]{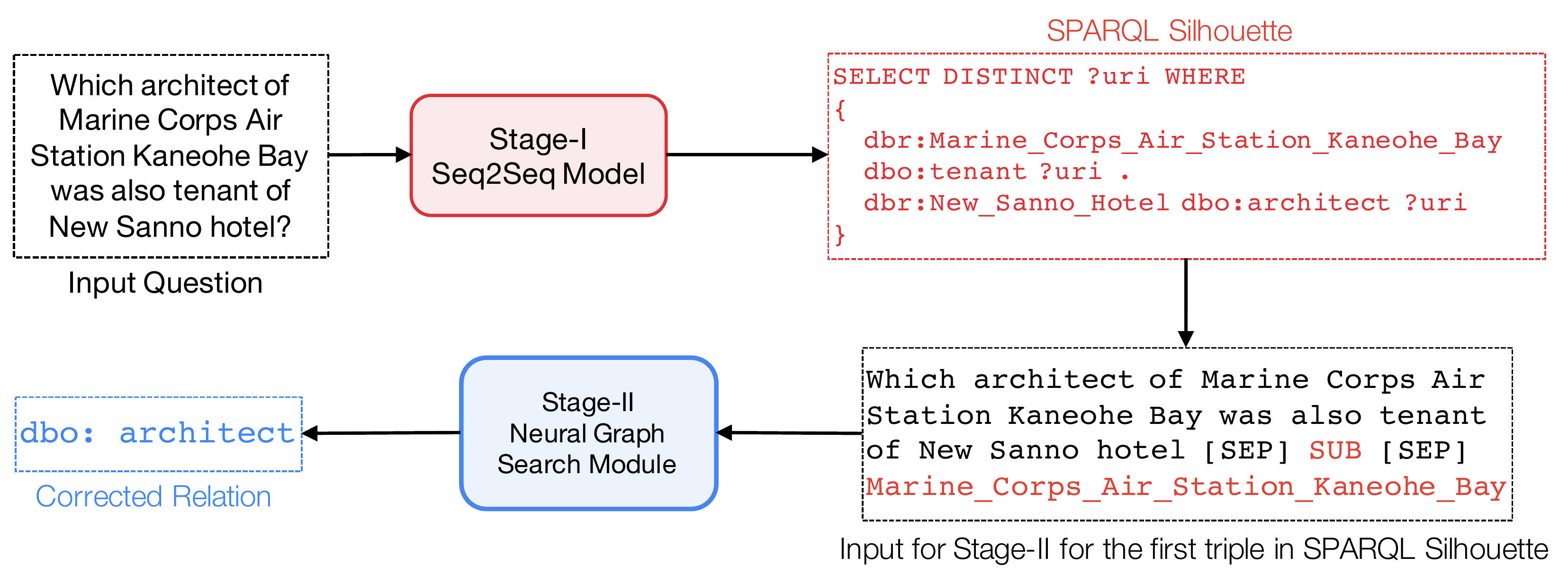}
\caption{An example of input to the neural graph search module.}
\label{input_graph_search}
\end{figure*}

\begin{table*}[!b]
\centering
\setlength\tabcolsep{8pt}
\renewcommand{\arraystretch}{1.8}
\begin{tabular}{lll}
\toprule
Question  &  \parbox{1.6cm}{Gold\\Relation}& \parbox{1.5cm}{Predicted Relation}\\
\midrule
\parbox{7.5cm}{Where was the person born who died in \textbf{Bryn Mawr Hospital}?}  & placeOfDeath & deathPlace \\
\midrule
\parbox{7.5cm}{ Name the rivers who originate from \textbf{Essex}?} & mouthPlace & sourceRegion\\
\midrule
\parbox{7.5cm}{Name the artist who made Dream Dancing and is often associated with \textbf{Joe Pass}.}
 & associatedBand   & associatedMusicalArtist\\
\midrule 
\parbox{7.5cm}{What is used as money for French Southern and Antarctic Lands is also the  product of the \textbf{Karafarin Bank} ?}	
 & product & products\\
\bottomrule
\end{tabular}
\caption{Anecdotal examples from LC-QuAD-1 test set where graph search module is unable to disambiguate between two closely related relations (gold and predicted) that are available for the highlighted entities in DBpedia.}
\label{failing_examples}
\end{table*}
\begin{table*}[h!]
\centering
\setlength\tabcolsep{9pt}
\renewcommand{\arraystretch}{1.2}
\begin{tabular}{lll}
\toprule
Question & Gold SPARQL & SPARQL silhouette\\
\midrule
\parbox{0.18\textwidth}{Who was called Scarface?} & 

\parbox{0.38\textwidth}{SELECT ?uri 
 WHERE \\ 
 \{\\
 ?uri dbo:alias ?alias\\
 FILTER \textbf{contains(lcase(?alias),\\"scarface")}\\
 \}} &
\parbox{0.33\textwidth}{SELECT DISTINCT ?uri WHERE\\ 
\{\\ 
\;\;dbr:Scarface dbo:alias ?uri\\
\}}\\
\midrule

\parbox{0.18\textwidth}{Who was called Rodzilla?}
& \parbox{0.38\textwidth}{SELECT DISTINCT ?uri WHERE\\
\{\\
?uri \textbf{\textless  http://xmlns.com/foaf/0.1/nick\textgreater \\
``Rodzilla''@en}\\
\}} &  
\parbox{0.33\textwidth}{SELECT DISTINCT ?uri WHERE \\ 
\{\\ 
dbr:Rodzilla dbo:alias ?uri \\
\}}\\
\midrule 
\parbox{0.18\textwidth}{Give me all gangsters from the prohibition era.} & 

\parbox{0.38\textwidth}{SELECT DISTINCT ?uri WHERE \\ 
 \{\\
 ?uri dbo:occupation dbr:Gangster ; \\ \textbf{dct:subject} \textbf{dbc:Prohibition-eragangsters}\\
 \}} &
\parbox{0.33\textwidth}{SELECT   DISTINCT   ?uri   WHERE\\ 
\{\\ 
?uri   a   dbo:Film  ; \\ 
dbo:time dbr:Gangsters\_of\_the\_Frontier\\ \}}\\
\bottomrule

\end{tabular}
\caption{Anecdotal examples from QALD-9 test set where gold SPARQL have a peculiar structure just because the specific way in which the corresponding facts are present in the DBpedia.}
\label{atypical_examples}
\end{table*}

\begin{table*}[h!]
\centering
\setlength\tabcolsep{7pt}
\renewcommand{\arraystretch}{1.2}
\begin{tabular}{lll}
\toprule
Question & Gold SPARQL & SPARQL silhouette\\
\midrule
\parbox{0.18\textwidth}{Which countries have more than ten volcanoes?}
& \parbox{0.38\textwidth}{SELECT DISTINCT ?uri WHERE \\
\{\\
?x a dbo:volcano ; \\ dbo:locatedInArea ?uri .\\ ?uri a dbo:Country\\
\} \textbf{GROUP BY} ?uri HAVING\\ ( COUNT(?x) \textgreater 10 )\\
} &  
\parbox{0.33\textwidth}{SELECT DISTINCT ?uri WHERE \\ 
\{\\ 
?uri a dbo:Country ;\\ dbo:location dbr:Countries\_of\_the\_United\_Kingdom \\
\}}\\
\midrule 
\parbox{0.18\textwidth}{Give me a list of all critically endangered birds.}
& 
\parbox{0.38\textwidth}{SELECT DISTINCT ?uri ?p WHERE \\
\{\\
?uri rdf:type dbo:Bird \\\{ ?uri dbo:conservationStatus "CR" \} \\ \textbf{UNION} \{ ?uri dct:subject \\ dbc:Critically\_endangered\_animals \}\\
\}} & 
\parbox{0.33\textwidth}{SELECT DISTINCT ?uri WHERE\\
\{\\
?uri a dbo:Film ; \\
dbo:principal dbr:Endangered\_Species\_(H.A.W.K.\_album \\
\}} \\
\midrule 

\parbox{0.18\textwidth}{Which daughters of British earls died at the same place they were born at?
} & 
\parbox{0.38\textwidth}{SELECT DISTINCT ?uri WHERE\\
\{\\
?uri rdf:type yago:WikicatDaughtersOfBritishEarls ; \\ dbo:birthPlace ?x ; \\ dbo:deathPlace ?y \textbf{FILTER} ( ?x = ?y ) \\
\}}  &
\parbox{0.33\textwidth}{SELECT DISTINCT ?uri WHERE\\
\{\\ 
 ?uri rdf:type yago:WikicatStatesOfTheUnitedStates ; \\
 dbo:place dbr:Daughters\_of\_the\_Dust\\
\}}\\
\bottomrule

\end{tabular}
\caption{Anecdotal examples from QALD-9 test set where gold SPARQL comprises infrequent SPARQL keywords. The corresponding SPARQL Silhouette predicted by our Stage-I is also shown for these examples.}
\label{missing_for_less_freq_keyword}
\end{table*}

\section{Evaluation Metric}\label{eval_met}

{\em 1) Precision, Recall, and $F_1$ for Single Question:} For single question $Q$, we compute precision $P$, recall $R$, and $F_1$ using the set of {\em gold answer entities} $S_g$ and {\em predicted answer entities} $S_p$. 
While computing these metrics, we handle boundary cases as follows. If $S_g=S_p=\emptyset$ then we take $P=R=F_1=1$. If only $S_g=\emptyset$ then we take $R=F_1=0$.\\
{\em 2) Macro Precision, Macro Recall, Macro $F_1$, and Macro $F_1$ QALD:}
These metrics are defined for the whole dataset. For this, we first compute $P$, $R$, and $F_1$ at individual question level and average of these numbers across entire dataset gives us the {\em macro} version of these metrics. For $F_1$, if use the boundary condition of having $P=1$ when $S_p=\emptyset, S_g\neq \emptyset$ then such a Macro $F_1$ is called as {\em Macro $F_1$ QALD} as per \citet{ngomo20189th}. But if we instead use $P=0$ then it is called Macro $F_1$.\\
{\em 3) Precision, Recall, and $F_1$ for the whole set:} For whole set, $P$ and $R$ are same as {\em macro} version of these metrics. $F_1$, however, is computed by taking Harmonic mean of these $P$ and $R$. The reported metrics for the {LC-QuAD-1} dataset were computed in this manner.\\ 
{\em 4) Answer Match (AM):} For a question $Q$, when executing the predicted SPARQL, if we have $S_p=S_g$ then we say AM=1 otherwise AM=0.\\

\section{Anecdotal Examples}\label{appen:anecdotal_examples}
Table \ref{failing_examples} shows examples from LC-QuAD-1 test set where our neural graph search module is unable to disambiguate between two very similar looking relations that exist in DBpedia for an entity.

Table \ref{atypical_examples} captures examples from QALD-9 test set where gold SPARQL have a peculiar structure just because the way in which corresponding facts are being captured within in the DBpedia and that makes it almost impossible for any KB agnostic techniques (such as seq2seq) to output such structures. The first two rows of Table \ref{failing_examples} shows examples where gold SPARQL queries of two very similar questions is quite different. Even though Falcon picks correct entities, our SPARQL silhouette struggle to yield two differently structured SPARQL queries for two very similar looking natural language questions. Third row of the table contains some entities/relations containing \textit{dct, dbc}, etc. Falcon linker does not tag these kinds of entity/relation, so we miss out correctly predicting the sketch in Stage-I and so in Stage-II as well. 

Table \ref{missing_for_less_freq_keyword} shows various examples from QALD-9 test set where we miss predicting the correct sketch of SPARQL because of very few number of such examples present in the training set. These are examples where SPARQL contains infrequent keywords such as GROUP BY, UNION, FILTER etc.
\end{document}